\documentclass[10pt, a4paper]{article}

\usepackage{lrec-coling2024} 
\usepackage{adjustbox}
\usepackage{multirow}
\usepackage{authblk}
\usepackage{hyperref}

\title{Large Language Models Are Also Good Prototypical Commonsense Reasoners}

\name{Chenglin Li$^{1,2\dagger}$, Qianglong Chen$^{1\dagger}$, Yin Zhang$^{1\ddagger}$, Yifei Zhang$^{2}$ Hongxiang Yao$^{3}$}

\address{\textsuperscript{1}Zhejiang University, \textsuperscript{2}Northeastern University, \textsuperscript{3}Ant Group
\\
\{22351307, chenqianglong, zhangyin98\}@zju.edu.cn, \\ zhangyifei@cse.neu.edu.cn, feiyu.fyyu@gmail.com
}

\abstract{
Commonsense reasoning is a pivotal skill for large language models, yet it presents persistent challenges in specific tasks requiring this competence. Traditional fine-tuning approaches can be resource-intensive and potentially compromise a model's generalization capacity. Furthermore, state-of-the-art language models like GPT-3.5 and Claude are primarily accessible through API calls, which makes fine-tuning models challenging. To address these challenges, we draw inspiration from the outputs of large models for tailored tasks and semi-automatically developed a set of novel prompts from several perspectives, including task-relevance, supportive evidence generation (e.g. chain-of-thought and knowledge), diverse path decoding to aid the model. Experimental results on ProtoQA dataset demonstrate that with better designed prompts we can achieve the new state-of-art(SOTA) on the ProtoQA leaderboard, improving the  Max Answer@1 score by 8\%, Max Incorrect@1 score by 4\% (breakthrough 50\% for the first time) compared to the previous SOTA model and  achieved an improvement on StrategyQA and CommonsenseQA2.0 (3\% and 1\%, respectively). 
Furthermore, with the generated Chain-of-Thought and knowledge, we can improve the interpretability of the model while also surpassing the previous SOTA models. We hope that our work can provide insight for the NLP community to develop better prompts and explore the potential of large language models for more complex reasoning tasks.
 \\ \newline \Keywords{Commonsense Reasoning, Large Language Model, Prompt}
 }

\begin{document}
\maketitleabstract

\section{Introduction}
Recently, large language models have made significant advancements in the field of natural language processing and have achieved SOTA results across a wide range of tasks~\citep{zhao2023survey,OpenAI2023GPT4TR,Bai2022ConstitutionalAH}.
The growth of model size plays a pivotal role in boosting their performance~\cite{zhao2023survey}. However, it is obvious that merely increasing the size of language models is insufficient to excel in tasks demanding intricate reasoning, particularly in the realm of commonsense reasoning~\cite{sap2020commonsense,Bhargava_Ng_2022}. Commonsense reasoning is the foundation of human understanding, rooted in the basic knowledge and life experiences accumulated through daily life and social practice~\cite{forguson1989common}, which outlines practical knowledge of how the world works~\cite{sap2020commonsense}. And commonsense reasoning is at the heart of building natural language understanding models that can reason about the world like humans do~\cite{davis2015commonsense,storks2019commonsense}. 

For the prototypical commonsense reasoning, the model output is expected to cover all prototypical answers for a question~\cite{boratko2020protoqa}. This is particularly valuable in scenarios where considering a given context should yield diverse and contextually appropriate responses~\cite{zhang2020task}. Meanwhile, in prototypical commonsense reasoning, traditional studies include building a ranker model~\cite{luo2022masked} and knowledge enhancement~\cite{li2023kepr} to improve the ability of the model from fine-tuning the model, but these works did not consider the commonsense reasoning ability of the SOTA models and limited to the specific fine-tuning tasks. ~\cite{bian2023chatgpt}have assessed large language models' knowledge related to prototype reasoning but utilized accuracy as the sole evaluation metric briefly, deviating from the original task's metrics.

In this work, we explore the prototypical commonsense reasoning ability of large language models and attempt to explain their reasoning process. Specifically, we draw inspiration from model responses to specific tasks and semi-automatically develop three distinct categories of prompts aimed at enhancing the model's performance.  

\begin{figure}
     \centering
    \resizebox{0.9\linewidth}{!}{\includegraphics{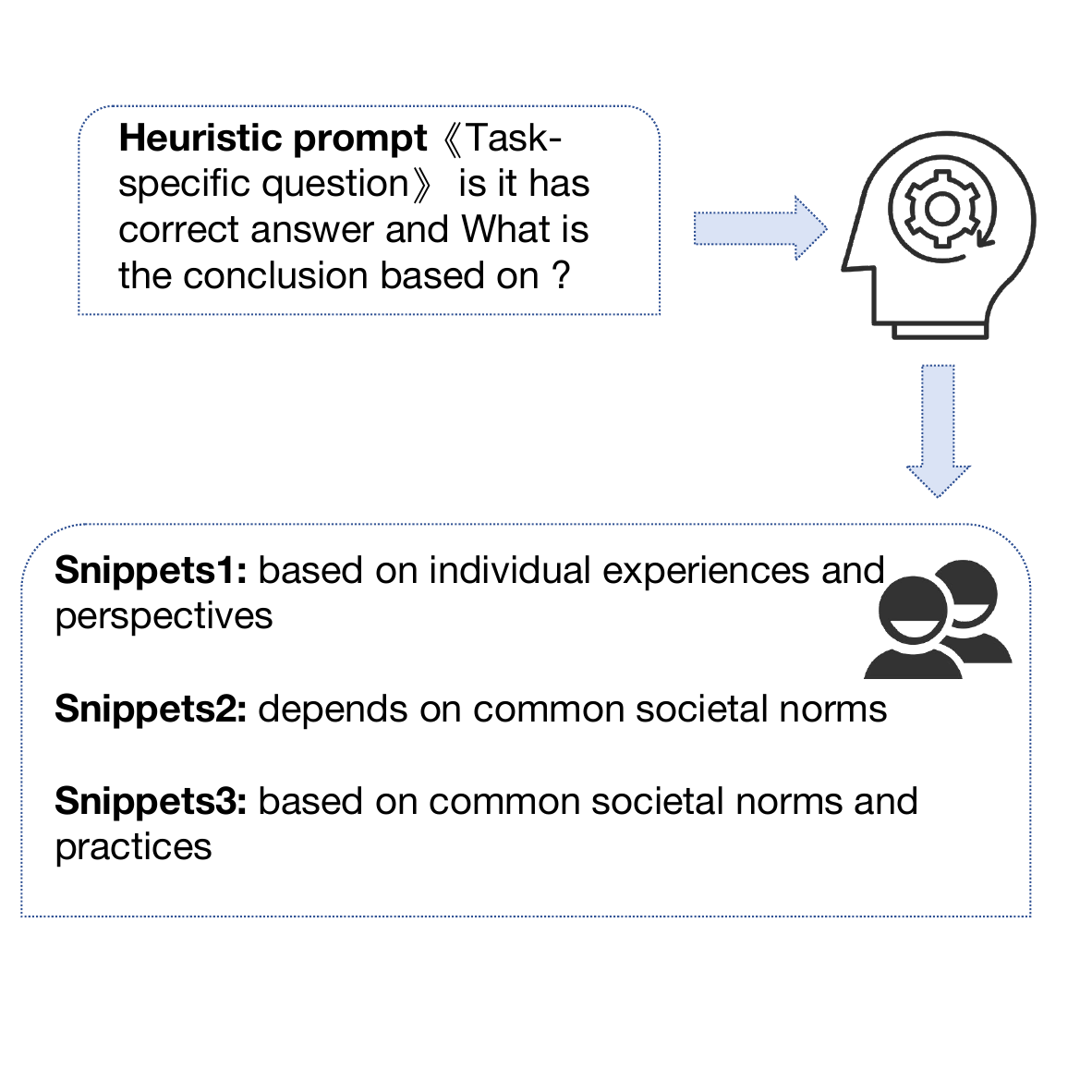}}
    \caption{Specific heuristic prompt we use \\ and snippets about ProtoQA task that we collect.}
    \label{fig:workflow-1}
\end{figure}

By semi-automatically developing three categories of prompts, we aim to gain a deeper understanding of These prompts encompass (1)the level of commonsense reasoning attainable by large language models in prototypical commonsense reasoning tasks, (2) an explanation of the commonsense reasoning processes employed by these models, and (3) strategies for enhancing the model's commonsense reasoning ability through improved new prompts.
Overall, we semi-automatically explore and develop a set of effective prompts for the prototypical commonsense reasoning from the perspectives of task-relevant, evidence-supported, and diverse path decoding. Our work contributes to a better understanding of large language models' capabilities in commonsense reasoning and offers insights into how to construct prompts that effectively elicit these abilities from such models.

The contributions of this work are as follows:
\begin{itemize}
    \item We semi-automatically explore a series of new prompts, including task-relevant prompt, evidence-supported prompt, and diverse path decoding prompt to improve the prototypical commonsense reasoning ability of large language models. 
    \item  To improve the interpretability of the model's reasoning process, we employ prompts based on chain-of-thought and knowledge generation, coupled with diverse path decoding prompts, thereby enhancing model performance while increasing transparency in the reasoning process.
    \item Experimental results demonstrate that our method achieve new SOTA on the ProtoQA. We improve the  Max Answer@1 score by 8\%, Max Incorrect@1 score by 4\% compared to the previous SOTA model. Meanwhile, we validate the generalization of our method on CommonsenseQA2.0 and StrategyQA.
\end{itemize}

\begin{figure*}
    \centering \includegraphics[width=1.0\linewidth]{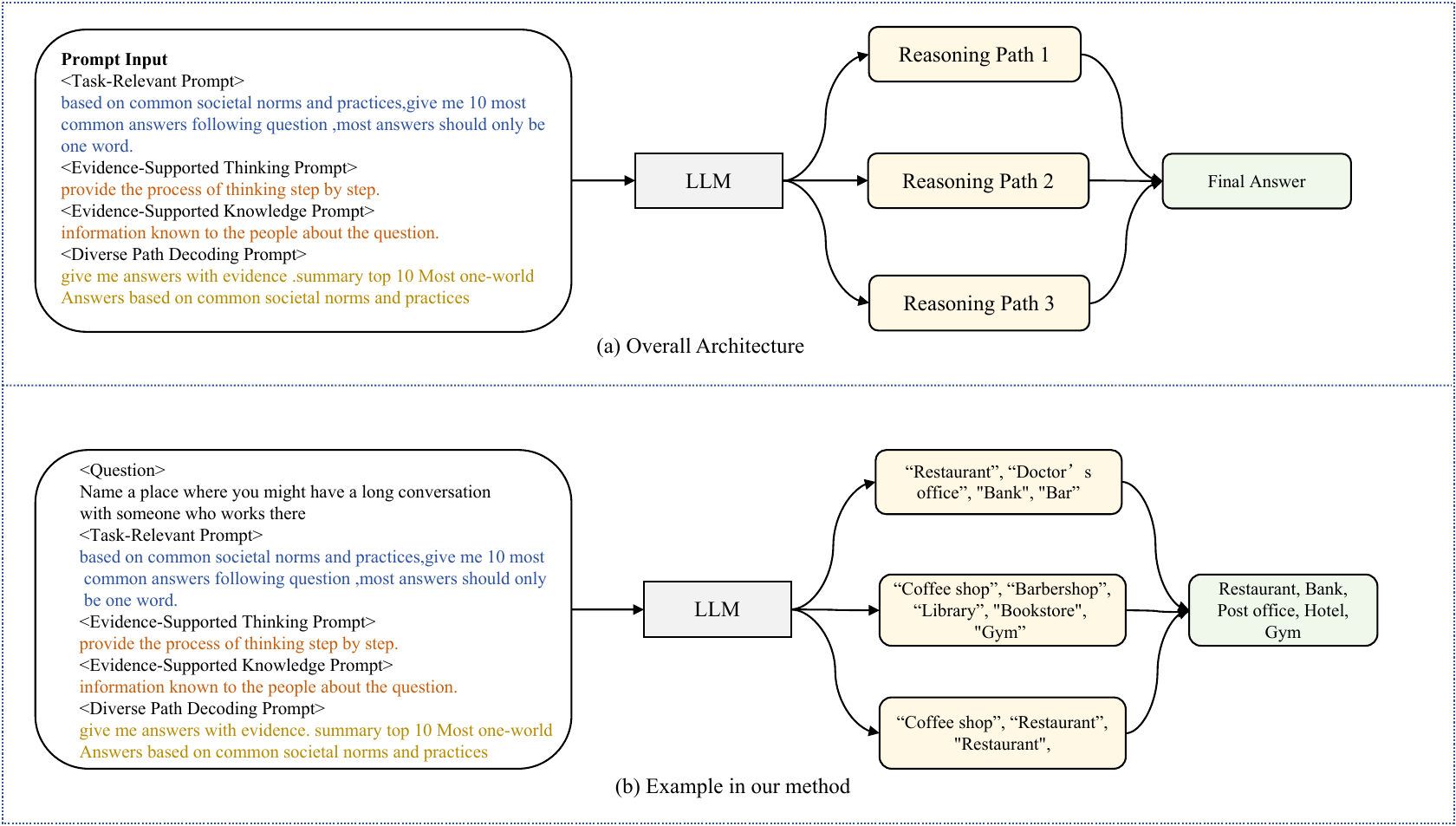}
    \caption{The overall architecture and example demonstration in our work. 
    }
    \label{fig:workflow-2}
\end{figure*}

\section{Related Work}
\subsection{Commonsense Reasoning with Large Language Models}
Large language models, such as GPT-3.5 series models~\footnote{https://platform.openai.com/docs/model-index-for-researchers}, GPT-4~\cite{OpenAI2023GPT4TR}, Claude~\cite{Bai2022ConstitutionalAH} and Google Bard~\footnote{blog.google/technology/ai/bard-google-ai-search-updates}, PaLM-2~\cite{Anil2023PaLM2T} have made great progress in the field of natural language processing. However, these large language models still encounter challenges in the realm of commonsense reasoning~\cite{bhargava2022commonsense}, where language model need have the capacity to reason implicit knowledge and daily situations that are known to humans~\cite{uleman2008spontaneous}. To enhance the common sense reasoning ability of large language models, researchers have explored diverse works which can be roughly categorized into how to make language model know more~\cite{yang2021survey} and how language model make better reasoning~\cite{wei2022chain}. In the former works, they integrate external sources of knowledge into language model~\cite{lim2020know,sun2019ernie,guan2020knowledge,yu2022survey,lin2019kagnet}.For instance, the utilization of external knowledge graphs or databases containing factual information can augment the model's understanding of the world~\cite{han2020open,liu2021kg}. By incorporating such knowledge, the model becomes capable of making more informed decisions~\cite{chen2020improving}. In the latter works, researchers design better reasoning strategies for the language model~\cite{kojima2022large}. By adopting an appropriate reasoning strategy, the ability of the language model to use knowledge reasoning is greatly improved~\cite{wang2022self}.
\subsection{Prompt Learning}
Prompt learning is a crucial process in the field of engineering, aimed at develop effective prompts to guide large language models in achieving desired outputs~\cite{white2023prompt}. Much work for prompt learning has been explored by researchers. A series of methods such as 
few-shot~\cite{wang2020generalizing}, Chain-of-Thought ~\cite{wei2022chain} in-context learning \cite{dong2022survey} essentially establishes suitable prompts to guide the language model to output the appropriate results. Among these methods, the Chain-of-Thought technology which generates thinking steps has been established as a significant method to enhance the reasoning ability of language models~\cite{wei2022chain}. Throughout various language tasks, prompt learning has played a pivotal role in achieving excellent results~\cite{dang2022prompt}. Currently, the research of NLP Prompt Engineering is still developing, and different methods have been proposed to construct effective prompts. Among them, the artificial construction of prompt is one of the most efficient methods, which is designed based on artificial knowledge~\cite{white2023prompt}.

\section{Approach}


In this section, we will introduce the semi-automatic prompting methods, which aims to elicit the common sense reasoning abilities of large language models.
Our main goal is to explore what properties a prompt possesses that can stimulate the commonsense reasoning ability of a large language model, and how a prompt can be used to balance the commonsense reasoning ability and interpretability of a large language model.

We mainly evaluate the performance of several large language models, including GPT-3.5, GPT-4, Claude, and Bard, on ProtoQA dataset~\citep{boratko2020protoqa}, StrategyQA~\citep{Geva2021DidAU} and CommonsenseQA2.0~\citep{talmor2021commonsenseqa}.

Inspired by the results of model when prompting model for specific tasks, we gather pertinent task-relevant snippets as an integral component of the prompts. Specific heuristic prompt we use and snippets about ProtoQA task we collect are shown in Figure~\ref{fig:workflow-1}. We select snippets3 that are closest to the task~\cite{boratko2020protoqa} as one part of the newly designed task-relevant prompts. While considering the model's interpretability and consistency, We separately designed  evidence-supported prompt and diverse path decoding prompt.


\begin{figure*}
    \centering
    \includegraphics[width=1\linewidth]{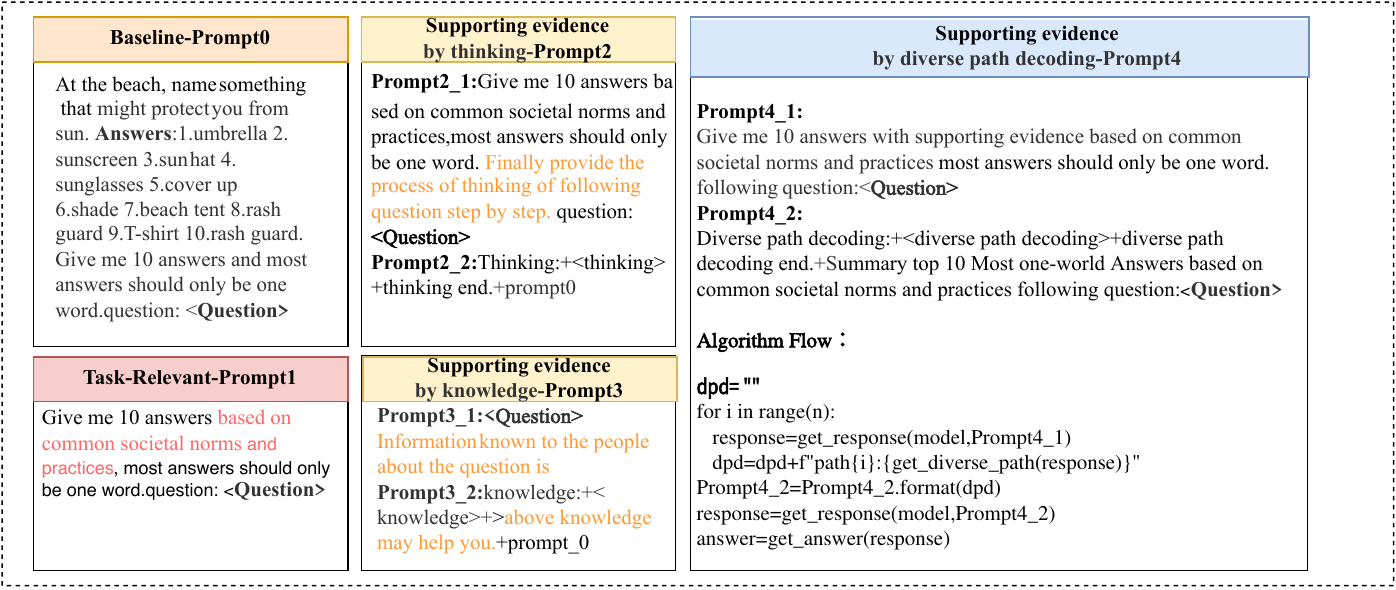}
    \caption{The prompt content in different setting with prompt label, where prompt0 denotes the baseline prompt, others denote different setting, such as task relevant, supporting evidence by thinking, knowledge and diverse path decoding. 
    }
    \label{fig:prompt_label}
\end{figure*}

\begin{itemize}
    \item \textbf{Task-Relevant} Adding task-relevant knowledge snippets  as prompts to generate diverse and accurate answers, the answer list needs to cover knowledge that is familiar to most people.We drew inspiration from the output of model, discovering their ability to generate task-relevant pieces. We then refined these pieces to craft task-specific prompts.
    
    Task-relevant prompts can focus the model's attention and produce output that is more in line with task requirements. At the same time, task-relevant prompts can provide contextual information for the model, helping the model understand the question and the type of output that needs to be generated. 
    \item \textbf{Evidence-Supported Thinking \& Knowledge} Adding instructive prompts  that include task-relevantguide words for models to generate supporting evidence, including Chain-of-Thought and knowledge generation, which can make the reasoning process of the model more transparent while the model generates diverse and accurate answers.

   Supportive evidence plays a significant role in various aspects. First, it enhances the transparency of the model, making the reasoning process and basis for judgment of the model clear and visible. At the same time, it can highlight the most important judgment information of the model and help humans understand the focus of the model. Furthermore, it supports manual inspection of the model's reasoning process, evaluates the rationality of the model, and improves the model.
    \item \textbf{Diverse Path Decoding} Considering the need for generating diverse and accurate answers, we add an instruction prompt that include task-relevant guide words, enabling the model to generate multi-dimensional answers in diverse reasoning process and relying on the model’s summarizing ability to produce the final result.
    Generate diverse answers and summary enables the model to attain enhanced self-consistency, which is crucial for the decision-making capabilities of the model. If contradictory or inconsistent output is generated, the model is highly likely to correct or discard it. 
\end{itemize}

As shown in Figure 2, we present the details of task-relevant prompt, which aims to guide the large language model to generate the answer with more task related information.
For the evidence-supported prompt, relevant knowledge or thought is first elicited through the prompt. 
The task-relevant prompt is then combined with evidence and input into the large language model to generate answers. 
For diverse path decoding prompt, multiple candidate answers with evidence are first decoded through multi-paths sampling. Then the candidate answer with evidence are summarized  by combining task-related prompt to generate the final answer. More details can be found in Appendix, as shown in Figure~\ref{fig:more_prompt}.

\section{Experiments}
\subsection{Dataset}
\paragraph{ProtoQA} In the main result, we evaluated our method on ProtoQA~\cite{boratko2020protoqa}, which focuses on commonsense reasoning questions such as "What could be some of the reasons you could be called to your kid's school?". The model is required to generate a list of answers, ideally covering all prototype answers for a given question. The dataset has 9,762 questions in the training set, 52 questions in the development set, and 102 questions in the test set. Datasets have diverse reasoning types of questions, as shown in Table 1.

\paragraph{CommonsenseQA2.0} CommonsenseQA 2.0~\cite{talmor2021commonsenseqa} is proposed to explore the commonsense understanding ability of large language models, which includes 14,343 yes/no questions (or assertions) about everyday commonsense knowledge.
\paragraph{StrategyQA}
StrategyQA~\cite{Geva2021DidAU} is a question-answering benchmark focusing on open-domain questions where the required reasoning steps are implicit in the question and should be inferred using a strategy, which includes 2,780 true/false examples, each consisting of a strategy question, its decomposition, and evidence paragraphs.

\begin{table}[]
 \centering
\begin{tabular}{|l|c|c|c|c|}
\hline
\textbf{Reasoning type}                & \textbf{Train}  & \textbf{Test} \\
\hline
Prototypical Events & 80\% & 68\%  \\
Habitual Activity   & 24\% & 40\%  \\
Event Reasoning     & 40\% & 28\%  \\
Specific Entities   & 4\% & 20\%   \\
Social/Mental       & 12\% & 16\%  \\
Negation            & 20\% & 12\%  \\
\hline
\textbf{Total numbers}   & 9,762 & 102  \\
\hline

\end{tabular}
 \caption{Statistics of different reasoning types on ProtoQA.}
\end{table}

\begin{table*}[]
\centering
 \adjustbox{width=1\textwidth}{
\begin{tabular}{|l|ccccccc|}
\hline
\multicolumn{1}{|l|}{\multirow{2}{*}{\textbf{Method}}}                           & \multicolumn{4}{c|}{\textbf{Max Answers}} & \multicolumn{3}{c|}{\textbf{Max Incorrect}} \\
\multicolumn{1}{|c|}{} &
  \textbf{@ 1} &
  \textbf{@ 3} &
  \textbf{@ 5} &
  \multicolumn{1}{c|}{\textbf{@ 10}} &
  \textbf{@ 1} &
  \textbf{@ 3} &
  \textbf{@ 5} \\ \hline
  GPT-2-KEPR      & 0.42 & 0.50 & 0.55 & 0.63 & 0.33 & 0.52 & 0.60 \\ \hline
BART-KEPR      & 0.47 & 0.56 & 0.57 & 0.65 & 0.38 & 0.55 & 0.62 \\ \hline
T5-3B-KEPR                                                             & 0.61   & 0.61   & 0.62  & \textbf{0.69}  & 0.46       & 0.61      & \textbf{0.68}      \\ \hline
T5-3B-finetune                                                         & 0.57   & 0.57   & 0.59  & 0.63  & 0.39       & 0.55      & 0.59      \\ \hline
GPT-3-davinci-5-Shot & 0.57   & 0.55   & 0.58  & 0.61  & 0.41       & 0.55      & 0.59      \\ \hline
GPT-3.5-few-shot & 0.62 & 0.63 & 0.62 & 0.64 & 0.47 & 0.60 & 0.62 \\ \hline
Claude-few-shot & 0.63 & 0.62 & 0.62 & 0.64 & 0.45 & 0.57 & 0.61 \\ \hline
Bard-few-shot   & 0.58 & 0.57 & 0.57 & 0.64 & 0.41 & 0.54 & 0.60 \\ \hline
GPT4-few-shot   & 0.64 & 0.61 & 0.62 & 0.66 & 0.48 & 0.58 & 0.62 \\ \hline

GPT-3.5-Task-relevant prompt & \textbf{0.69}   & \textbf{0.66}   & \textbf{0.66}  & 0.68  & \textbf{0.50}       & \textbf{0.63}      & 0.66      \\ \hline
GPT-3.5-Support evidence thinking prompt &
  0.66 &
  0.65 &
  0.65 &
  0.68 &
  0.47 &
  0.62 &
  0.64 \\ \hline
  GPT-3.5-diverse path decoding prompt &
  0.67 &
  0.64 &
  0.65 &
  0.67 &
  0.48 &
  0.62 &
  0.64 \\ \hline
\end{tabular}
}
\caption{Test results on ProtoQA leaderboard, compared with other baselines, our method achieves new SOTA.}
\label{leaderboard}
\end{table*}

\subsection{Baselines}
For the test set experiment, we take other SOTA models on the leaderboard as baselines for comparison as shown in Table ~\ref{leaderboard}.  
For the development set experiment, the baseline of large language models is the standard few-shot prompt, where the model is provided contextual examples of input-output pairs before predicting the output on a test example. Additionally, given the brevity and diversity of the an swers of the question, we impose a limit in the prompt stating "give me 10 answers and most answers should only be one word.", as shown in Figure 3. The following introduces the models in the experiment which have achieved excellent performance and large language models that we explore.
\bigbreak
\paragraph{GPT-2} GPT-2 is a transformer-decoder-based language model developed by Open AI\cite{radford2019language}, which demonstrates strong performance on a variety of language modeling tasks like next word prediction, sentence completion, and text generation. We use the model fine-tuned on the training set.
\paragraph{BART}
BART is a Transformer-based language generation model developed by FaceBook AI with a bidirectional encoder and decoder~\cite{lewis2019bart} which enables it to simultaneously understand the context of the input text and generate the output text, allowing it to generate more coherent and accurate responses. We use the model fine-tuned on the training set.
\paragraph{T5} T5 is a powerful multi-task learning language model based on the transformer architecture with encoder and decoder developed by Google AI~\cite{raffel2020exploring}, which realizes multi-task learning and transfer learning by unifying different natural language processing tasks into text-to-text conversion. We use the model fine-tuned on the training set.
\paragraph{KEPR} GPT-2-KEPR, BART-KEPR, T5-3B-KEPR utilizes Knowledge Enhancement and Plausibility Ranking to augment the commonsense reasoning skills of GPT-2, BART and T5~\cite{li2023kepr}. We use the model fine-tuned on the training set. 
\paragraph{GPT-3} GPT-3 is the updated version of GPT, a large AI language model with 175 billion parameters which shows powerful language abilities in text generation, The GPT-3 version in use is the davinci-002 model released by OpenAI. ~\cite{brown2020language}. 
\paragraph{GPT-3.5} GPT-3.5 is the updated version of GPT, which make greater process in dialogue system~\cite{ouyang2022training}. The GPT-3.5 version in use is the GPT-3.5-turbo model released by OpenAI.
\paragraph{Bard} Bard is an experiment based on LaMDA which is trained to be informative and comprehensive~\cite{thoppilan2022lamda}.
\paragraph{Claude} Claude is a next-generation AI assistant based on Anthropic’s research, which can help with use cases including summarization, search, creative and collaborative writing, Q\&A, coding, and more~\cite{bai2022training}.

\begin{table*}
    \centering
    \adjustbox{width=0.7\textwidth}{
   \begin{tabular}{|l|ccccccc|}
\hline
\multirow{2}{*}{\textbf{Prompt label}} & \multicolumn{4}{c|}{\textbf{Max Answers}}                                                & \multicolumn{3}{c|}{\textbf{Max Incorrect}}         \\
                              & \textbf{@ 1} & \textbf{@ 3} & \textbf{@ 5} & \multicolumn{1}{c|}{\textbf{@ 10}} & \textbf{@ 1} & \textbf{@ 3} & \textbf{@ 5} \\ \hline
Prompt0 & 0.55 & 0.50 & 0.47 & 0.48 & 0.32 & 0.40  & 0.45 \\ \hline
Prompt1 & \textbf{0.70} & \textbf{0.57} & 0.53 & 0.54 & \textbf{0.39} & 0.46 & 0.50 \\ \hline
Prompt2 & 0.55 & 0.57 & 0.53 & 0.53 & 0.39 & 0.47  & 0.50 \\ \hline
Prompt3 & 0.56 & 0.52 & 0.48 & 0.50 & 0.36 & 0.43  & 0.46 \\ \hline
Prompt4 & 0.54 & 0.54 & \textbf{0.55} & \textbf{0.59} & 0.38 & \textbf{0.49} & \textbf{0.54} \\ \hline
\end{tabular}
    }
\caption{Results with different prompts on ProtoQA development set.}
 \label{dev_result}
\end{table*}

\subsection{Experimental Setting}

\paragraph{Task Relevant Prompt}
The main dataset we studied is the ProtoQA 
dataset, which requires the generation of diverse and accurate answers based on common societal norms and practices, so we choose the piece 'based on common societal norms and practices' as the key part of the task-relevant prompt, as shown in Figure 3.

\paragraph{Prompt that generates supporting evidence}
We consider generating supportive evidence through 
Chain-of-Thought or knowledge generation. We first elicit thinking or knowledge through an initial prompt, then use generated thought or knowledge in conjunction with the task-relevant prompt as input to generate answers, as shown in Figure 3.

\paragraph{Prompt that helps diverse path decoding}
For the model to generate answers with evidence multiple times, finally summarize to generate the final answers, as shown in Figure 3. 
In the parameter of calling GPT-3.5-turbo model API, $\mathrm{temperature}=0.5$, $\mathrm{max\_tokens}=1024$, $\mathrm{top\_p}=0.95$.

\subsection{Experimental Results}
We use ProtoQA, the commonsense reasoning dataset, to evaluate the performance of language model with different prompt, which provides a more rigorous assessment of the commonsense reasoning ability of large language models, which requires the model to output all prototype answers to a question ideally. The metric Max Answers@k limits the total number of answers allowed to up to k answers. The metric Max Incorrect@k allows unlimited answers but stops after k unmatched answers. Therefore, the metrics evaluate the diversity and accuracy of the answers to a given question. The higher the score, the better the common sense reasoning performance. All results presented here are the averages derived from multiple runs, specifically the mean of three repetitions, and the results hold statistical significance.

\subsubsection{Results on Leaderboard of Test set }




In the leaderboard of test set, as shown in Table~\ref{leaderboard}, we achieved SOTA results through a task-relevant prompt which outperforms various models enhanced by KEPR method. Specifically, the metric Max Answers@1 reached 69\% (an 8\% increase over the previous SOTA model, T5-3B-KEPR), and the metric Max Incorrect@1 reached 50\%. This is the first time breaking through 50\%, representing a 4\% increase over the previous SOTA model. Furthermore, to explore the commonsense reasoning capabilities of other large language models, we conducted experiments with Claude, Bard, and GPT-3.5 using few-shot prompts, as shown in Table~\ref{leaderboard}.
Through the large language models with few-shot prompts, the metrics Max Answers@10 and Max Incorrect@5 are behind the previous SOTA models. 
This indicates that with knowledge enhancement and ranker techniques, smaller language models have the potential to match or even outperform larger language models on some specific tasks. Additionally, the large language model, Claude, demonstrated commonsense reasoning ability comparable to that of GPT-3.5, while maintaining a clear advantage over Bard.

\begin{figure*}
    \centering
    \includegraphics[width=0.90\linewidth]{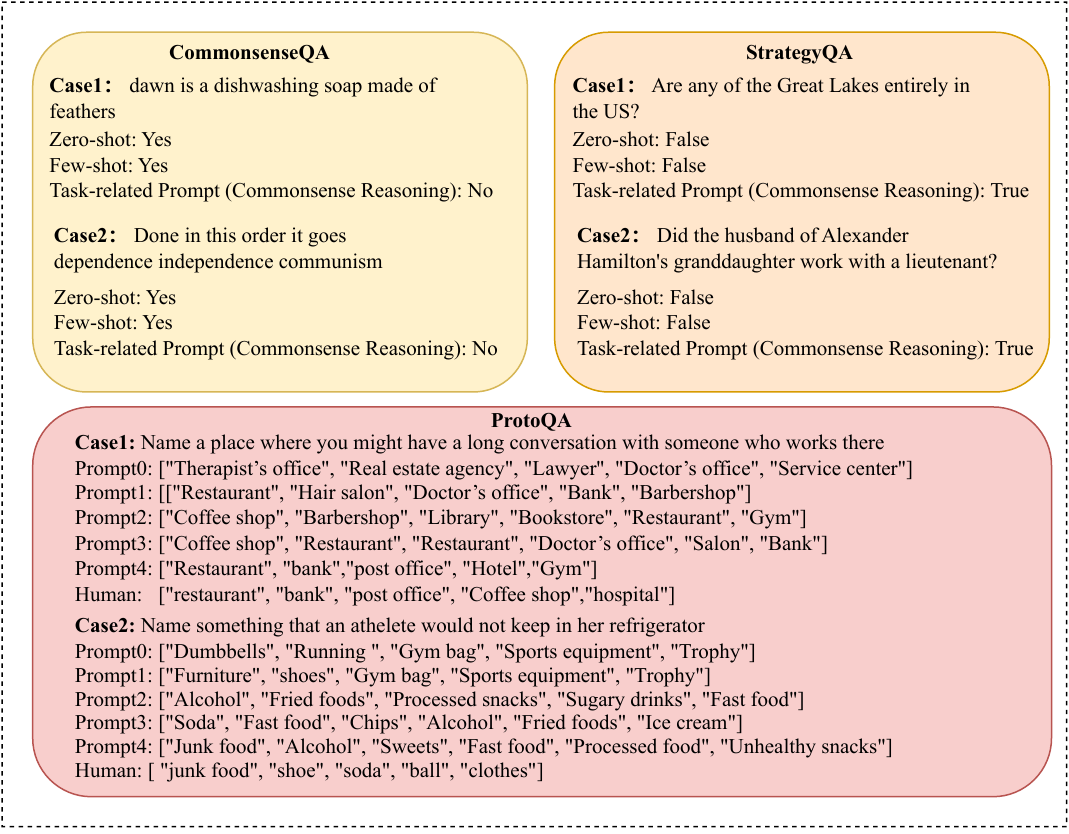}
    \caption{Case study on ProtoQA, CommonsenseQA and StrategyQA}
    \label{fig:case_study}
\end{figure*}

\subsubsection{More Results on Development set}

As shown in Table~\ref{dev_result}, experimental results demonstrate that task-relevant prompt (Prompt1) substantially contribute to the performance of model. Specifically, the metrics Max Answers@1, an increase of 15\% and Max Incorrect@1 an increase of 7\%. In addition, when providing reasoning explanations aims to improve the interpretability of the model (Prompt2 and Prompt3), its commonsense reasoning competence has not deteriorated under most evaluation metrics. For the metric Max Answer@1, the model falls back to the baseline reasoning level. However, for other metrics, the model have a significant improvement compared with the baseline (Prompt0).
Furthermore, through the diverse path decoding prompt(Prompt4), the reasoning ability of the model has been significantly improved. In particular, the metric Max Answer@10 has increased by 11\%, and the metric Max Incorrect@5 has increased by 9 \% compared with the baseline. 

\begin{table}
    \centering
    \resizebox{\linewidth}{!}{
    \begin{tabular}{|l|c|c|}
    \hline
    \textbf{Model Setting} & \textbf{StrategyQA} & \textbf{CommonsenseQA2.0} \\
    \hline
    Few-shot & 0.585 & 0.57 \\
    \hline
    Task-Relevant Prompt & 0.594  & 0.60 \\
     \hline
    Evidence-Supported Thinking & 0.576  &  \textbf{0.655}\\
     \hline
    Evidence-Supported Knowledge & 0.603  &  0.650 \\
     \hline
    Diverse Path Decoding & \textbf{0.607}  &  0.65 \\
    \hline
    
    \end{tabular}
    }
    \caption{Performance on other datasets, LLM we used is GPT-3.5-turbo.}
     \label{generalization-1}
\end{table}

\begin{table}
    \centering
    \resizebox{\linewidth}{!}{
    \begin{tabular}{|l|c|c|}
    \hline
    \textbf{Model Setting} & \textbf{StrategyQA} & \textbf{CommonsenseQA2.0} \\
    \hline
    Few-shot & 0.52 & 0.50 \\
    \hline
    Task-Relevant Prompt & 0.55  & \textbf{0.55} \\
     \hline
    Evidence-Supported Thinking & 0.55  &  0.51 \\
     \hline
    Evidence-Supported Knowledge & 0.54  &  0.50 \\
     \hline
    Diverse Path Decoding & \textbf{0.57}  &  0.54 \\
    \hline
    
    \end{tabular}
    }
    \caption{Performance on other datasets, LLM we used is LLAMA-70B.}
     \label{generalization-2}
\end{table}
    
\begin{table*}
    \centering
    \adjustbox{width=0.7\textwidth}{
   \begin{tabular}{|l|ccccccc|}
\hline
\multirow{2}{*}{\textbf{Prompt label}} & \multicolumn{4}{c|}{\textbf{Max Answers}}                                                & \multicolumn{3}{c|}{\textbf{Max Incorrect}}         \\
                              & \textbf{@ 1} & \textbf{@ 3} & \textbf{@ 5} & \multicolumn{1}{c|}{\textbf{@ 10}} & \textbf{@ 1} & \textbf{@ 3} & \textbf{@ 5} \\ \hline
Prompt0 & 0.34 & 0.31 & 0.34 & 0.37 & 0.18 & 0.28  & 0.32 \\ \hline
Prompt1 & \textbf{0.51} & \textbf{0.40} & \textbf{0.40} & 0.42 & \textbf{0.27} & 0.33 & 0.38 \\ \hline
Prompt2 & 0.38 &  \textbf{0.40} & 0.39 & 0.42 & 0.22 & 0.33  & 0.38 \\ \hline
Prompt3 & 0.37 & 0.36 & 0.37 & 0.40 & 0.20 & 0.30  & 0.35 \\ \hline
Prompt4 & 0.46 & 0.39 & \textbf{0.40} & \textbf{0.43} & 0.24 & \textbf{0.34} & \textbf{0.40} \\ \hline
\end{tabular}
    }
\caption{Performance on LLAMA-70B in ProtoQA development set}
 \label{dev_result}
\end{table*}

\subsubsection{Generalization Evaluation}
To demonstrate the effectiveness of our method, we conduct experiments on other commonsense reasoning datasets, CommonsenseQA2.0 and StrategyQA where each question need a single output, as shown in Table~\ref{generalization-1} and Table~\ref{generalization-2}. 
By adding task-relevant prompt, we achieve an improvement of about 3\% on  CommonsenseQA2.0 and about 1\% on StrategyQA. By adding evidence-supported thinking, we achieve the best performance on CommonsensenQA2.0, which gains 8.5\% improvements.
For prompts of CommonsenseQA2.0 and StrategyQA, we all used the same feature of prompts (Based on social common sense).

\subsection{Ablation Study}
 The task-relevant prompt (Prompt1) method achieves a significant improvement in the commonsense reasoning ability of the large language model.
 In the ProtoQA development set, the metric Max Answer@1 is increased by 15\% compared to the baseline with few-shot prompt (Prompt0), as shown in Table~\ref{dev_result}. In the ProtoQA test set, the task-relevant prompt (Prompt1) helps the large model improve by 7\% compared to GPT3.5 with few-shot prompt, as shown in Table~\ref{leaderboard}. There are about 3\% improvement in CommonsenseQA2.0 and 1\% improvement in StrategyQA, as shown in Table~\ref{generalization-1}. From the above experimental results, we observe that task-relevant prompts assist the model in achieving excellent performance on particular tasks, similar to fine-tuning language models on specific tasks to enhance their capabilities for specific tasks. At the same time, when we provide explanation for the model reasoning process, the model's common sense reasoning ability keeps competitive when k>=3. In the ProtoQA development set, providing supportive evidence with thinking process (Prompt2) improves the Max Incorrect@3 metric by 7\% compared to the baseline(Prompt0). Providing supportive evidence with knowledge (Prompt3) improves the Max Incorrect @3 metric by 3\%, as shown in Table~\ref{dev_result}. In the ProtoQA test set, providing supportive evidence with thinking improves the Max Answer@10 metric by 4\% compared to GPT3.5 with the few-shot prompt, as shown in Table~\ref{leaderboard}. The experimental results demonstrate that the augmentation of model interpretability leads to outputs that are more comprehensive. However, We found that the evidence-supported prompt will have a certain loss in the performance of the model. One factor that contributed to the performance degradation is its excessive focus on the cognitive process, which introduced redundant information.

Additionally, through employing diverse path decoding prompt, the commonsense reasoning ability of the large language model has also been significantly improved which relieve this loss when incorporating supportive evidence prompt. In the ProtoQA development set, the metric Max Answers@10 increased by 11\% and the metric Max Incorrect@5 increased by 9\% compared to the baseline(Prompt0), as shown in Table~\ref{dev_result}. The results of the above experiment demonstrate that the reasoning ability of the model can be enhanced through diverse path decoding. Specifically,the large language model is able to produce more precise outputs by summarizing the candidate answers and maintain competitiveness in metrics with k>=3.

\subsection{Case study}

To elaborate on the prompts, we provide the actual output of the large language model with different prompts for commonsense reasoning task, as shown Figure~\ref{fig:case_study}. 
We can observe that through task-relevant prompts, the large language model produces the correct output and improves the accuracy of answers to CommonsenseQA2.0 and StrategyQA questions. For ProtoQA dataset, compared with few-shot Prompt0, the answer list output by the model appears more reasonable with task-relevant Prompt1, which enables model to generate coherent and relevant answers while reducing the possibility of errors. Specifically, there are no obvious irrational answers (e.g. "real estate agency", "lawyer" for the first question, "Name a place where you might have a long conversation" is irrational. "running" for the second question,"Name something that an athelete would not keep in her refrigerator" is irrational). 

Meanwhile, the supportive evidence generated by the model itself (process of thinking Prompt2 or knowledge Prompt3) enables it to gain a better understanding of the question and generate multiple accurate answers  (e.g. "coffee shop" for the  first question. "fast food", "alcohol" for the second question). In our artificial evaluation of Cot, we noted a complete alignment between Cot's reasoning process and its ultimate outcomes.We hypothesize that a substantial enhancement of the model's reasoning capabilities by CoT might potentially lead to an increased occurrence of disparities between CoT's reasoning process and its decision-making outcomes. In our artificial evaluation of related knowledge, we found that knowledge can well supplement the background information of the problem and help the model understand the question. The supportive evidence of detailed output are provided in the Appendix \ref{evidence}.
Additionally, by diverse path decoding Prompt4, the model output is closer to the question. (e.g. "junk food" for the second question).This detailed output are provided in the Appendix \ref{decoding}.
In addition, we found that there were differences in the performance of the model between the test set and the development set, which was caused by the different difficulty levels of the dataset distribution. The overall difficulty of the development set was higher than that of the test set.

\section{Conclusion}
In this paper, we explore the prototypical commonsense reasoning capability of large language models to generate multiple appropriate outputs.
We semi-automatically developed a set of novel
prompt templates as a simple and broadly applicable method to enhance commonsense reasoning in large language models. Our research findings indicate that large language models exhibit competitive commonsense reasoning abilities without the need for external knowledge enhancement. In fact, incorporating explicit external knowledge into these models may introduce redundant information, potentially compromising their performance. However, by employing diverse decoding strategies, we can mitigate such performance loss while simultaneously enhancing the interpretability of the models.  In experiments on the ProtoQA, StrategyQA, and CommonsenseQA datasets, we found that incorporating task-relevant knowledge can enable models to reduce the likelihood of error and generate more accurate answers. By facilitating thought process and knowledge, the model's faculty for commonsense reasoning can be more transparent. Through the diverse path decoding, the common sense reasoning ability of the model has also been improved to a certain extent and maintain competitiveness compared to evidence-supported prompt with single path decoding.

\section{Limitations}
Our work illustrates the task-relevant prompt, evidence-supported prompt and diverse path decoding prompt can improve the commonsense reasoning ability of large language models. However, due to the resource limitation, we are unable to conduct experiments on advanced GPT-4. Meanwhile, many reasoning datasets for evaluating language reasoning ability focus on measuring a model's ability to produce a single correct or logical response. We are limited to the reasoning datasets with multiple plausible outputs. Specifically, we  carry out experiments on the ProtoQA, a multi-output dataset, and conduct expanded experiments on the single-output CommonsenseQA and StrategyQA datasets. We hope more datasets that evaluate language models' ability to generate multiple appropriate outputs will be constructed, as this capability of large language model is crucial.

\bibliographystyle{lrec-coling2024-natbib}
\bibliography{lrec-coling2024-example}


\clearpage
\newpage
\appendix
\label{sec:appendix}
\begin{figure*}
    \centering
    \includegraphics[width=1\linewidth]{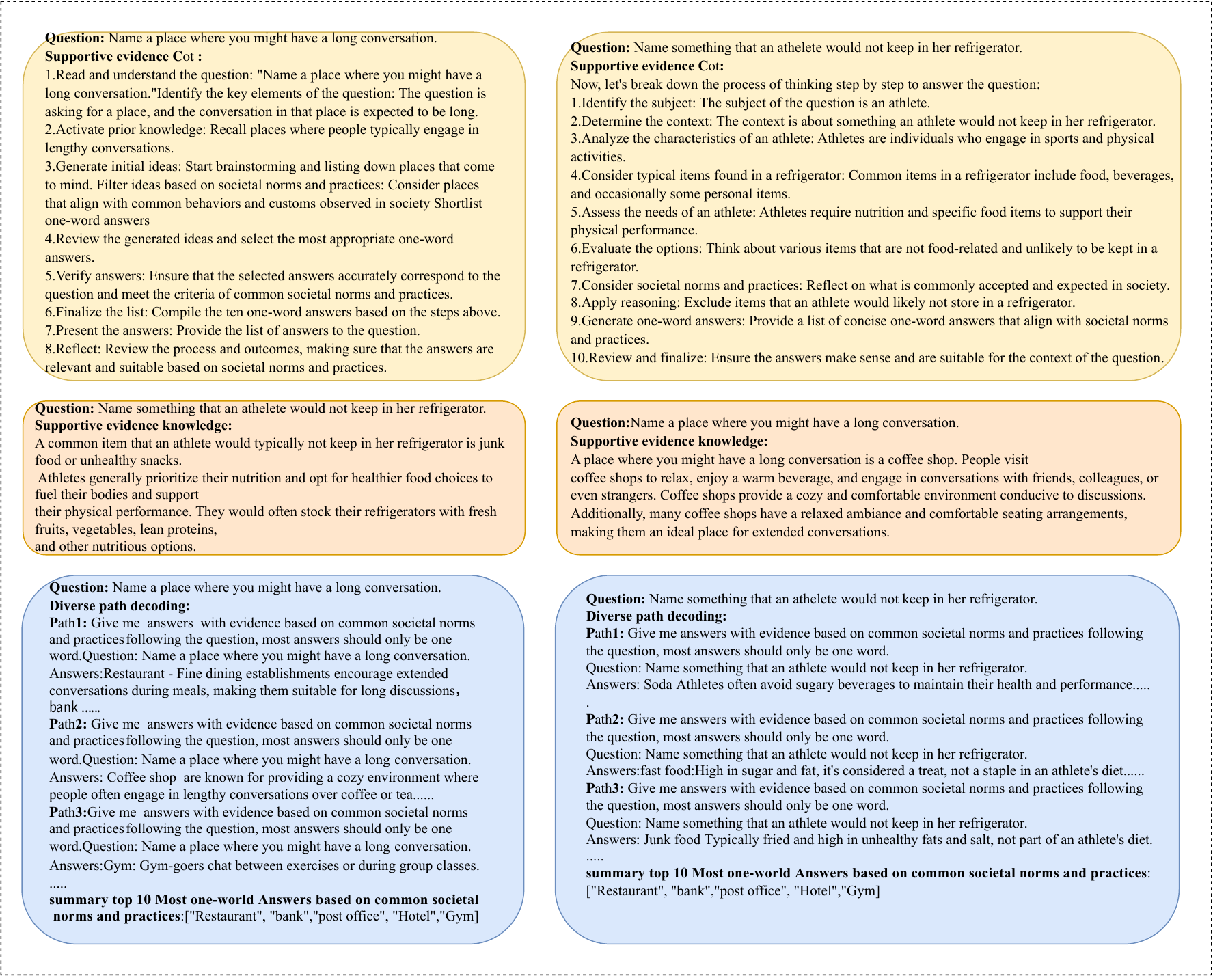}
    \caption{The details of supportive evidence thinking, supportive evidence knowledge and diverse path decoding.}
    \label{fig:more_prompt}
\end{figure*}

\section{Appendix: Details of prompt}

\subsection{Details of supportive evidence prompt}
\label{evidence}
As shown in Figure~\ref{fig:more_prompt}, for the question "Name a place where you might have a long conversation", the generated supporting thought evidence shows the reasoning process of the model on answering the question. It identifies the key  of the question, 'a place' and suggests having a long conversation at that place. The supporting knowledge evidence shows the reference knowledge of the model to answer the question, and provide the background knowledge of the gold answer 'coffee shop'. For the question,"Name something that an athelete would not keep in her refrigerator", model also produces the similar output. In short, through evidence-supported prompts, it generates reasonable output and improves the interpretability of the model's reasoning.
\subsection{Details of diverse path  decoding prompt}
As shown in Figure~\ref{fig:more_prompt}, in response to the question,"Name a place where you might have a long conversation", the model generated three times results and eventually filtered out inappropriate output such as "Patio" and "Beach" through summarization and sampling.For the question "Name something that an athelete would not keep in her refrigerator", the model generates  outputs that are more typical  through summarization and sampling, such as "junk food" or "alcohol". Overall, through diverse path decoding prompt, the outputs of the model are more accurate and diverse.
\label{decoding}

\end{document}